# Nearest Neighbor Classification for Classical Image Upsampling

Evan Matthews*
evanmm3@illinois.edu

Nicolas Prate*
nprate2@illinois.edu

*University of Illinois Urbana-Champaign

## Abstract

Given a set of ordered pixel data in the form of an image, our goal is to perform upsampling on the data such that: the resulting resolution is improved by some factor; the final result passes the 'human test' having added new, believable, and realistic information and detail to the image; and the time complexity for upsampling is relatively close to that of lossy upsampling implementations.

We are informally collaborating with Mark Bauer (markb5@illinois.edu) and Quinn Ouyang (qouyang3@illinois.edu) to accomplish the same problem but with different approaches. In contrast to Mark and Quinn's learning-based approach to intelligently upsampling given prior knowledge [1], we have taken a more classical approach in which we attempt to upsample given only direct information from the input.

## 1. Metrics and Testing

We generate and compile images for testing from a set of images containing a variety of styles, scapes, colors, and patterns. For each image, multiple test images are generated by downsampling the original by various factors. Our methods are subsequently applied to each, upsampling back to the original image. Provided both the original image and these upsampled reconstructions, we apply metrics to measure similarity, including RMSE, PSNR, and SSIM. By leveraging these metrics, we are able to gain differing perspectives and insights into the nature of differences between original images and our reconstructions, providing means to better visualize the strengths and weaknesses of our approaches.

### 1.1. Baseline: Linear-Interpolation

In addition to measuring the effectiveness of our approaches, we used linear-interpolation upsampling from the module *opencv-python* as a baseline. This standard approach is tasked with upsampling our test images as described above, with hopes that our methods will outperform these baseline capabilities.

### 1.2. MSE, RMSE, MAE

Fundamentally, we consider raw pixel comparison with the Root Mean Squared Error (RMSE) and Mean Absolute Error (MAE)- variations of Mean Squared Error (MSE)- which calculate the root mean square/absolute differences between pixels of the original image and our upsampled reconstruction, respectively. While a theoretically ideal upsampling implementation would return an exact copy of the original image, we only expect our results to *minimize* this metric, as a true recovery of lost information is impossible in the classical sense.

### 1.3. PSNR

The Peak Signal to Noise Ratio (PSNR) measures the ratio between the maximum value of an image and the power of corrupting noise, with higher values indicating a higher-quality reconstruction. This metric allows us to make larger structural comparisons between the base and upsampled images produced through our experiment.

### 1.4. SSIM

The Structural Similarity Index Measure (SSIM) measures differences in structural information, texture, and luminance, with higher values indicating a higher-quality reconstruction. The maximum value of '1' represents a SSIM between identical images. Similar to PSNR, we will be using SSIM to make comparisons regarding the structural integrity of upsampled images rather than individual pixel comparisons. Between the previously stated metrics, SSIM is most likely to align with human perception.

## 2. Methods and Approaches

We considered three approaches as improvements over linear-interpolation-based upsampling. First, we implemented K-Nearest-Neighbors (KNN) based upsampling as a means of scaling the resolution of fixed-aspect-ratio images by an integer factor. Next, we expanded our KNN approach to support upsampling with dynamic aspect ratios: upsampling to some ratio differing from the original image while maintaining the structure of the image. Finally, we further expanded the approach to support selective upsampling: saving runtime by optimizing over image regions of constant or near-constant color.

## 2.1. KNN-Interpolation

Our baseline is the simplest expansion over linear interpolation to solve the problem of image upsampling - filling in new pixels with the average of its k-nearest neighboring pixel values. Additionally, for the baseline approach, we support only fixed aspect-ratios between original and upsampled images, which equates the baseline functionality to 'deblurring' an image as other nearest-neighbor implementations have expected [2].

Despite being a baseline approach, it still posed some interesting implementation challenges. Depending on the size of the input image and desired size of the output, there are a varying number of pixels in various patterns requiring interpolation. For example, if the desired output size is reasonably larger than the input, there will be 'seas' of empty pixels that need filling; with too small a K, there will be no original pixels in the neighborhood to consider for interpolation. Additionally, interpolation around the corners and borders of the image provide a smaller neighborhood to work with, and thus a less-informed estimation will occur.

In our implementation, a new image is created with dimensions of the original image multiplied by some factor. Next, we loop through each pixel in this new image. If both pixel indices (i, j) are evenly divisible by the factor, we know this pixel position should be a pixel directly copied from the original image. If this is not the case, we know we must apply KNN interpolation. This interpolation will average all pixels from the original image that are within 'k' indices from the current pixel.

## 2.2. Upsampling for Dynamic Aspect Ratios

Our first expansion allows upsampling to aspect ratios differing from that of the original image, bringing new challenges and opportunities. For example, when upsampling one dimension while leaving the other unchanged, it would make sense to interpolate using pixels only along the changing dimension, rather than considering the true k-neighborhood as the baseline does.

Expanding the functionality to support dynamic aspect ratios alone was not too challenging, involving only adding two factor parameters, one for each dimension, rather than just one as in the base KNN-interpolation, and ensuring each factor is used with calculations relating to its respective dimension.

This generalization of the problem raises a more general issue of which nearest-neighbors to choose for pixel prediction. In a sense, we would ideally not want to rely on a constant aspect ratio when making predictions about neighboring pixels. For example, an image whose height remains constant but width doubles should ideally interpolate on horizontal nearest-neighbors and stretch the image while maintaining its vertical content. This elaboration also increases robustness compared to the baseline, allowing ideal upsampling for odd-width or odd-height images and the ability to upsample images to resolutions not accessible to the original aspect ratio.

## 2.3. Selective upsampling

In the sense of time complexity, more rigorous upsampling algorithms take more time to handle unknown pixels, continually getting worse for higher-resolution inputs and larger outputs. As such, selective upsampling was our attempt to drastically cut down on runtime. For instance, if our input was an image of a face on a perfect, solid-colored background, we would expect selective upsampling to focus most of the runtime on upsampling the face, being able to avoid repeatedly calculating the same pixel value with KNN interpolation across the background. While this elaboration may not be practical for all images, we expect selective upsampling to work reasonably well on real-life photos and realistic images - inputs where large, solid-color regions or regions with noticeable color patterns are most present. Additionally, we were inspired to pursue this extension of our implementation from research on similar image problems in which certain "features" were found to reduce time complexity [3].

This, of course, would demand an increase in overhead and complexity to detect regions where selective upsampling could be appropriate in an image, but our belief is that this overhead will be outweighed by the runtime saved in KNN calculations. We leverage edge detection and color-gradient calculations to assist in the distinction between regions with more 'constant' pixel values and more detailed regions requiring KNN interpolation.

Additionally, it is important to tune the size of region required to trigger selective upsampling functionality; too small a threshold may result in grainy or pathy results with our KNN interpolation applied nowhere, while too large a threshold may result in the entire image being upsampled with KNN interpolation in addition to the extra overhead we introduced in attempt to detect regions.

It is thus an invaluable goal to find some threshold that both decreases runtime on applicable images and maintains the quality produced by previous approaches.

Our implementation identifies pixels in a color region by deciding whether neighbors of a given pixel is within a certain threshold of the established color value. If this is true, then our implementation considers the pixel to be in a region, reducing the runtime of the image interpolation by using existing color values- as opposed to computing near-similar KNN interpolated values.

## 3. Results

For a classical non-learning based approach, our results were more than satisfactory, as even our baseline KNN-interpolation was able to upsample images with reasonable detail.

### 3.1. KNN-Interpolation

For its relative simplicity, the baseline approach yielded great results with minimal undesirable artifacts. Pitfalls of this approach were most visible surrounding detailed regions of crisp color change, such as borders between objects where our approach produced grainy transitions not present in the original image. Additionally, since this approach must perform a KNN average for every new pixel, runtime increases exponentially with output sizes.

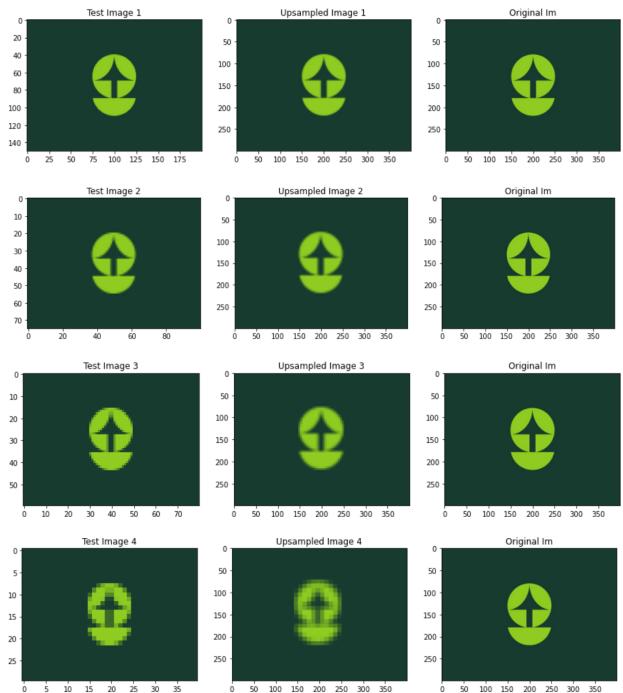

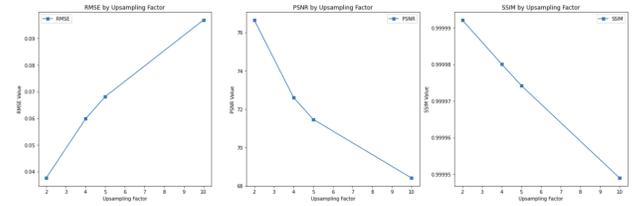

The above diagrams depict an image of a logo downsampled by various factors (2, 4, 5, and 10), then upsampled back to original dimension with our base KNN-interpolation algorithm. Looking at the images themselves- particularly the last row- the averaging behavior of our KNN-interpolation is very clear, with the upsampled image almost appearing as a blurred version of the original. Comparing these upsamplings to the test images they came from, however, the added information does make the logo object appear more defined; the edges are more contoured like the original, rather than blocky as in the test images. Looking at the plots of RMSE, PSNR, and SSIM, it is clear that the larger the factor used to downsample, (then upsample), the larger the error compared with the original image. Specifically, RMSE and PSNR appear inversely correlated, growing and falling in similar proportions. SSIM steadily decreases as the upsample factor increases. This makes sense, as the larger a factor used to downsample, the more information is lost and required to be estimated by our algorithm.

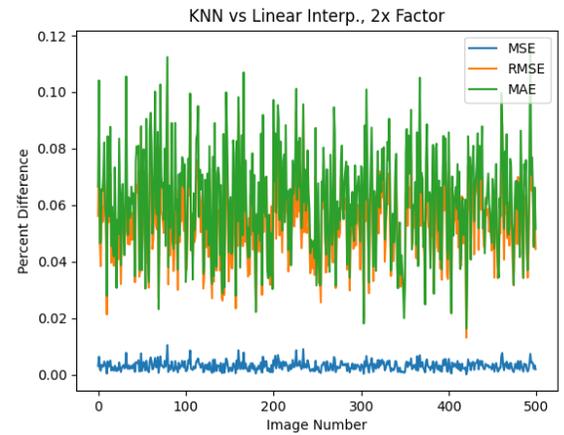

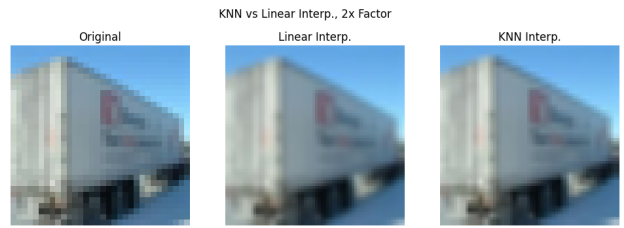

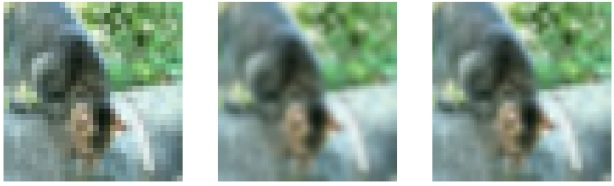
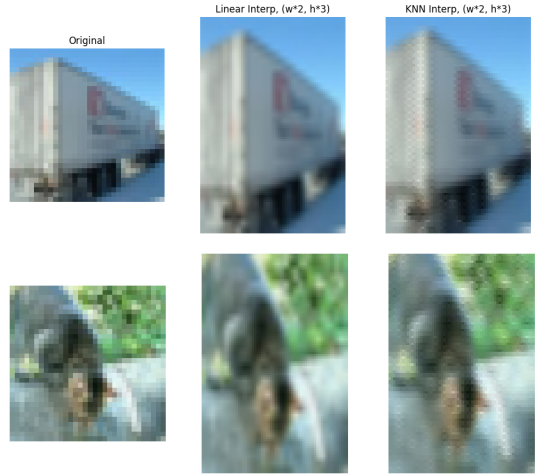

Across a subset of the CIFAR-10 dataset [4], upsampled by a factor of 2, RMSE and MAE show a promising range of roughly 1-12% error between our implementation and linear interpolation. Visually, KNN interpolation succeeds our expectations as it maintains the structural quality of the original image, despite a lack of edge-direction as implemented by other algorithms [5][6]. Given the percent differences encountered, however, we suspect that these metrics are an accumulation of mildly distinguishable interpolation differences across the entire image.

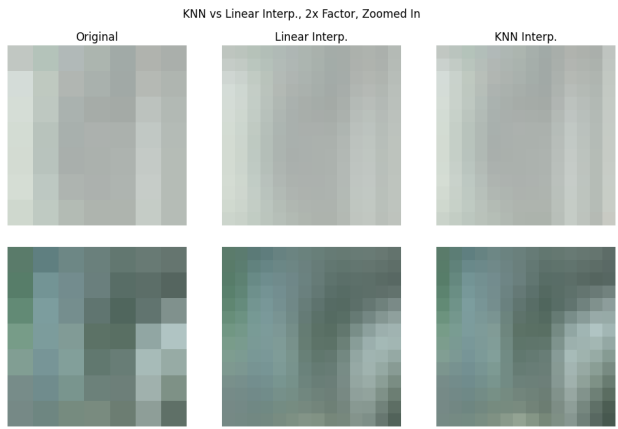

We can best observe this suspicion by viewing smaller subsets of our results, such as in the following subplots where the most prominent colors from any discernible shapes differ by about a couple pixels in location.

### 3.2. Upsampling for Dynamic Aspect Ratios

Dynamic Aspect Ratio Upsampling was accomplished by generalizing our existing KNN-interpolation code. In the case of single-dimensional upsampling, the algorithm is easily modified to consider only pixels along the dimension in question during interpolation. This result is achieved by scaling and interpolating pixel values with respect to the original width and height, with one dimension scaled by a positive integer factor. This process scales to upsampling with multi-dimensional upsampling where the dimension-scaling factors are not equal; since our implementation checks for pixels requiring interpolation with respect to any new image size, any two unequal, positive integer factors can be used to upsample the image.

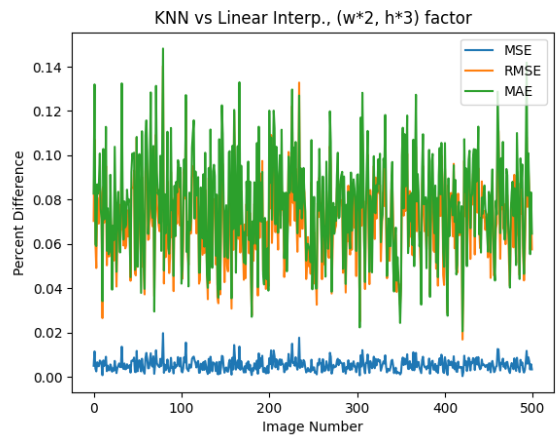

### 3.3. Selective Upsampling

Selective upsampling, our final expansion on the KNN implementation, provided a couple logistical challenges during development. Initially, our plan was to adapt the range of nearest neighbors relative to content from an edge-analysis plot. Ideally, edges detected on an image would require separate calculation since edge blurring is a common issue among most interpolation algorithms. Considering that we idealized selective upsampling to improve the runtime on our KNN implementation, we instead chose to focus solely on regions of constant or near-constant color.

We have demonstrated below three scenarios using this color-gradient selective upscaling approach: a more ideal image applying selective upscaling only on pixels with zero color-gradient (grad_thresh=1), a less ideal image applying selective upscaling only on pixels with zero color-gradient (grad_thresh=1), and the same less ideal image applying selective upscaling on pixels with a gradient less than 20 on a 0-255 scale (grad_thres=20)..

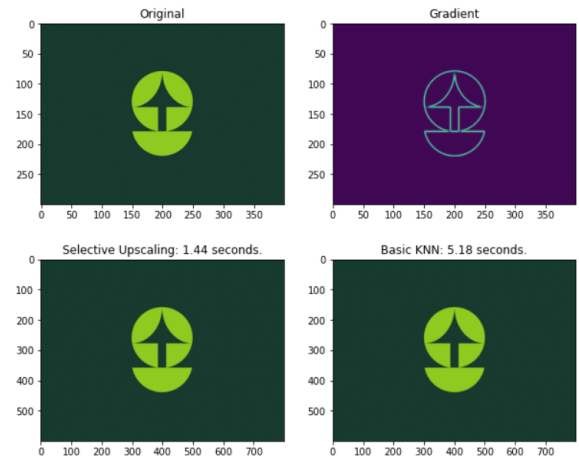

Above, we see for a more ideal image, one with a small number of constant colors, selective upscaling is able to produce practically identical results compared to the base KNN implementation at a fraction of the runtime while only applying selective upscaling to pixels with zero gradient.

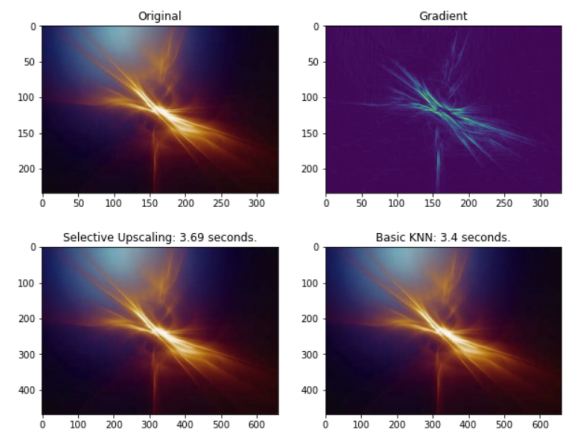

Above, we see for a less ideal image, one with a large number of changing colors, selective upscaling is able to produce practically identical results compared to the base KNN implementation while only applying selective upscaling to pixels with zero gradient, but takes longer than the base KNN implementation. This is because the number of pixels with zero gradient is small enough that we don't save computational time to overcome the extra overhead of selective upscaling.

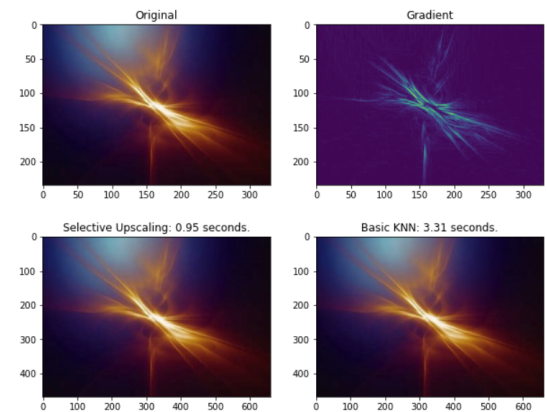

Finally, above we see that for the same lessideal image, selective upscaling is able to produce visually identical results (see SSIM) compared to the base KNN implementation- at a fraction of the runtime- while applying selective upscaling to pixels with gradient less than 20. By increasing the gradient threshold with which selective upscaling is triggered, we force our algorithm to interpolate less and copy more, even when the gradient is not zero. As a whole, this parameter and example shows that selective upscaling has the potential to drastically cut down runtime of upsampling compared to the base KNN approach- while maintaining visual coherence- even on images that are theoretically not ideal for such optimization.

## 4. Conclusions

By delving into a classical approach for image upsampling, we have shown that data generation without prior knowledge or learning can relatively match machine learning models with notably less resources. We believe that our results, even at a relatively visible difference in accuracy, provides a few benefits: a massive decrease in storage costs for upsampling tools, an added 'robustness' in terms of what images can be upsampled, and a potential decrease in computation time needed to generate the upsampled image.

## 5. Acknowledgements

We would like to personally thank Dr. Paris Smaragdis for providing us the opportunity to pursue this research endeavor, as well as Mark Bauer and Quinn Ouyang for their collaboration in the topic of image upsampling. Additionally, we would like to thank YouTube personality BobbyBroccoli, as without his video "The image you can't submit to journals anymore," we may have fallen into the trap of using the "Lena" picture in an image processing paper.